\documentclass[letterpaper, 10 pt, conference]{ieeeconf}  

\usepackage{balance}
\usepackage{graphicx}
\usepackage[font=small,skip=5pt]{caption}

\IEEEoverridecommandlockouts                              

\usepackage{graphicx}
\usepackage{cite}

\usepackage{verbatim}

\usepackage[dvipsnames]{xcolor}

\title{\LARGE \bf
An Attention Transfer Model for Human-Assisted Failure Avoidance in Robot Manipulations
}

\author{Boyi Song$^{+}$, Yuntao Peng$^{+}$, Ruijiao Luo$^{+}$, Rui Liu$^{*}
$
\thanks{Authors are with the Cognitive Robotics and AI Lab (CRAI), College of Aeronautics and Engineering,
        Kent State University, Kent, OH 44240, USA. $^{+}$ denotes that the first three authors have equal contributions to this paper. $^{*}$ Rui Liu is the corresponding author ruiliu.robotics@gmail.com }%
}
\usepackage{balance}  

\begin{document}

\maketitle
\thispagestyle{empty}
\pagestyle{empty}


\begin{abstract}
Due to real-world dynamics and hardware uncertainty, robots inevitably fail in task executions, resulting in undesired or even dangerous executions. In order to avoid failures and improve robot performance, it is critical to identify and correct abnormal robot executions at an early stage. However, due to limited reasoning capability and knowledge storage, it is challenging for robots to self-diagnose and -correct their own abnormality in both planning and executing. To improve robot self diagnosis capability, in this research a novel human-to-robot attention transfer (\textit{\textbf{H2R-AT}}) method was developed to identify robot manipulation errors by leveraging human instructions. \textit{\textbf{H2R-AT}} was developed by fusing attention mapping mechanism into a novel stacked neural networks model, transferring human verbal attention into robot visual attention. With the attention transfer, a robot understands \textit{what} and \textit{where} human concerns are to identify and correct abnormal manipulations. Two representative task scenarios: ``serve water for a human in a kitchen" and ``pick up a defective gear in a factory" were designed in a simulation framework CRAIhri with abnormal robot manipulations; and $252$ volunteers were recruited to provide about 12000 verbal reminders to learn and test \textit{\textbf{H2R-AT}}. The method effectiveness was validated by the high accuracy of $73.68\%$ in transferring attention, and the high accuracy of $66.86\%$ in avoiding grasping failures.
\end{abstract}


\section{Introduction}
Influenced by real-world dynamics and hardware uncertainty, robots inevitably fail in task executions. 
Robot abnormal behaviors result in various hazards, including economic loss, threats to human safety, and decreased social acceptance of robots. 

Failure avoidance is an urgent need for improving robot performance \cite{c28} \cite{c29}, yet it is a challenging practice in the real world. First, it is hard for a robot to realize that its performance is abnormal \cite{c27}. Accurate and prompt failure detection is difficult due to the high requirements for both advanced sensing systems and reasoning algorithms. It is challenging to design a reasoning system that both plans task executions and simultaneously monitors execution abnormalities \cite{c30}\cite{c31}. Moreover, even when a robot can realize its abnormalities, it is difficult for it to identify the abnormal executions and correct them correspondingly \cite{c32}\cite{c64}. Lastly, it is expensive to correct robot failures. The extra perceiving, reasoning, and action systems increase costs of robot system design and deployment \cite{c90}.

\begin{figure}[!t]
  \centering
 \includegraphics [width=0.92 \linewidth ]{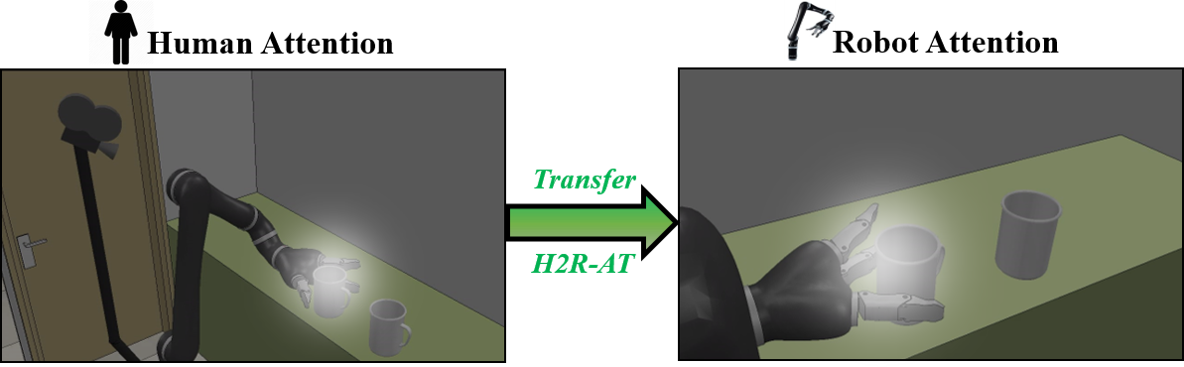}
  \caption{An illustration of the attention transfer using the developed \textit{\textbf{H2R-AT}}  model. The attention region of a human (from human observation perspective) and robot (from robot perceiving perspective) are highlighted as shown. By using \textit{\textbf{H2R-AT}} the attention of abnormal robot executions was transferred from a human to a robot to alert its failures in an early stage before failures happen.}
  \label{illustration}
  \vspace{-0.6cm}
\end{figure}

\begin{figure*}[ht!]
  \centering
 \includegraphics [width=0.8 \linewidth ]{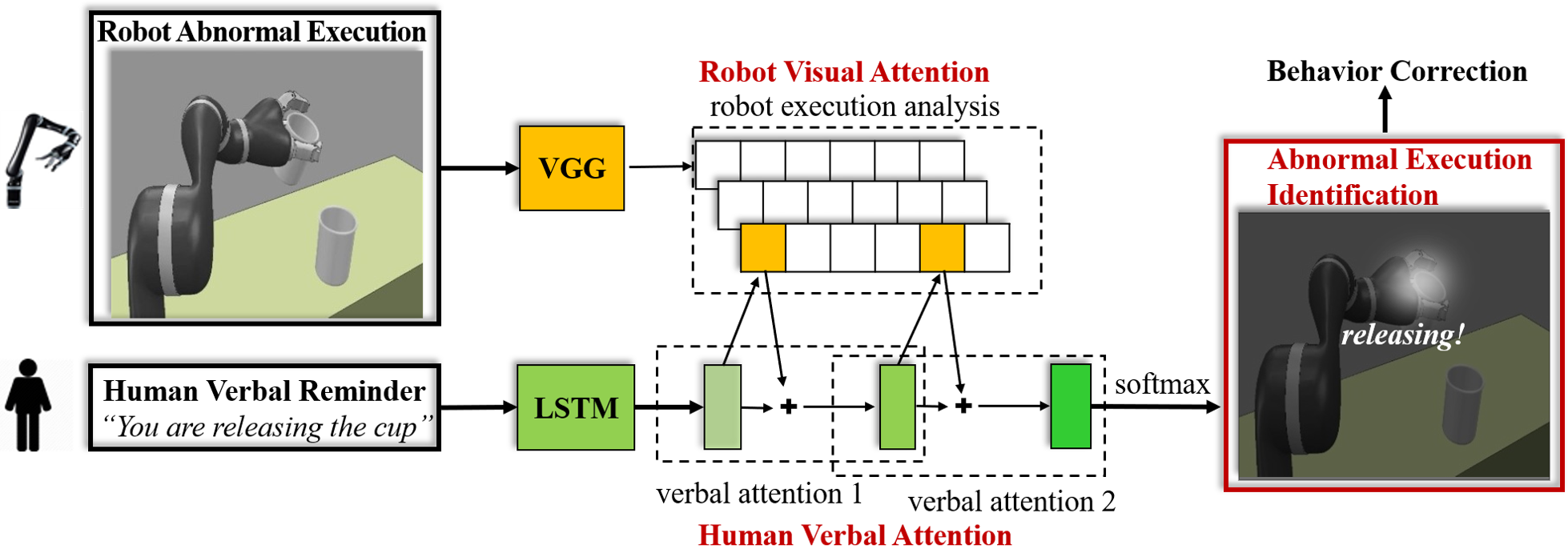}
  \caption{The framework of \textbf{\textit{H2R-AT}} using human verbal reminders for robot failure avoidance. Human attention is embedded in verbal reminders. Feature vectors extracted from human verbal reminders and feature vectors from robot visual perceiving are combined to get confined attention. With the confined attention, the robot can correct its abnormal behavior accordingly.}
  \label{H2R-AT}
  \vspace{-1em}
\end{figure*}

To address these challenges, a novel human-to-robot attention transfer (\textbf{\textit{H2R-AT}}) method, as shown in Figure \ref{illustration}, was developed in this paper, by introducing human intelligence to detect abnormal robot executions at an early stage and subsequently correct executions to avoid failure. Human attention is reflected in the concern of the specific area of their perceiving. When abnormal behaviors occur, human address their concern immediately on the fault area and generate the possible reason based on their domain knowledge. Through \textbf{\textit{H2R-AT}}, human intent is transferred to robots and help them perceive the abnormal execution. In this research, we envision a human-guided robotic system. Human monitors robot execution and verbally alerts robots to the abnormal executions. The research made two contributions:
\begin{itemize}
\item { A novel attention transfer method was developed to transfer human attention of unsatisfied robot executions to a robot, to alert a robot of its abnormality.}
\item {Developed an attention supported failure correction method to help with the identification of robot abnormal executions for performance improvement.}
\end{itemize}

\section{Related Work}
Attention sharing was widely investigated in the robotics field to indicate human preference and clarify robot confusion \cite{p6}\cite{p7}. The attention mechanism was used in both daily and industrial scenarios to increase robot execution efficiency, improve robot execution accuracy, ensure human safety, and increase robot social acceptance \cite{c34}\cite{c38}\cite{c63}. For example, a social robot used human-like gestures according to human attention in a conversation to increase human engagement in interactions\cite{p3}\cite{c37}\cite{c35}; a service robot changed its trajectory by estimating human intended places to avoid collision with the human\cite{e1}\cite{p4}; an industrial robot followed human head orientations to find the intended place to improve object search and delivery accuracy \cite{c40}\cite{c39}. Even though attention mechanisms have been used in robotics research, there is minimal work focusing on robot failure avoidance. The \textit{\textbf{H2R-AT}} presented in this paper targeted failure avoidance by utilizing an attention mechanism to involved in a human to send timely alerts for abnormal robot behaviors.

Current attention transfer methods require prior user training which is expensive and time-consuming. Non-verbal attention was used to express human expectations to guide robot executions. Safety concern attention was delivered by using human gaze to indicate the cared human location to avoid collision \cite{c45}\cite{c47}. Social etiquette attention was delivered by recognizing facial expressions to suggest human willingness to cooperate \cite{c41}\cite{c42}\cite{c43}. Human preference attention was delivered by using hand gestures to point to the human-desired personal items for daily assistance \cite{c46}\cite{c49}. 
Though non-verbal attention is effective in delivering human instructions to robots, due to the needs in adding extra perceiving devices and reasoning algorithms, such as computer vision systems and image intelligence methods, to extract human instructions, non-verbal attention is expensive in designing robotic systems. Also, non-verbal attentions allow for only limited interaction patterns, restricting the content in human instructions sent to a robot and further limiting the implementation scope of robotic systems. In this work, the proposed \textbf{\textit{H2R-AT}} enables a robot to directly process human verbal instructions with an accurate understanding, supporting natural human guidance on robot failure avoidance with no requirements for prior user training, complex vision or sensor systems, thus reducing the cost for the robot failure avoidance.

\section{Attention Transfer Model}

When abnormal executions occur, the human will give a verbal alert to correct robot behaviors. The transfer of attention can help the robot to understand human alert and correspondingly identify abnormal robot executions by localizing human attention regions onto robot perceived actions.

As shown in Figure \ref{H2R-AT}, by using verbal reminders, human attention to suspicious robot behaviors is expressed. Based on Stacked Attention Networks \cite{san}, we designed a new model, \textbf{\textit{H2R-AT}}, combined with analysis methods in human verbal reminder processing and visual feature extraction of abnormal robot executions. The first-layer attention generated by combining these two factors are then multiplied to the robot perceiving, added to the human reminder feature to be the new reminder input for the second-layer attention. Human attention has been correlated with specific regions in robot perceiving, which is directly correlated with some robot executions, for finally identifying the abnormal robot executions according to human attention.

\subsection{Interpreting Human Intention from Verbal Reminders}
 
Human verbal alerts described the location and types of robot abnormal executions. Based on a Long-Short Term Memory (LSTM) model, a model suitable for sequential input, commonly used for linguistic data, the semantic meaning embedded in human verbal alerts can be extracted.

The Natural Language Processing (NLP) module can identify different reminders (concerns) accurately because of the use of LSTM and word embedding. LSTM has a strong temporal modeling capability in extracting meaning from temporal human verbal instruction which is suitable for dynamic scenarios where a human gives a continuous description.

The NLP module used LSTM instead of other semantic analysis methods because the human reminders do not have a fixed length and usually vary both in the form and in the meaning. Using LSTM means less training and a better accuracy. The human reminder is usually short, less than 15 words. This also makes LSTM suitable.

Considering a human natural language reminder $r = [ r_1 , r_2 , ... r_I ]$ where $I$ represents the length of the reminder and $r_i$ represents a ``one-hot" vector of the $i^{st}$ word of the reminder.

Let $M_{we}$ represent the word embedding matrix, which can show the robot the relations of different words. The matrix is used to convert the words to vectors $M_i$.  
\begin{equation}\label{eq:1}
M_i = M_{we} \cdot r_i ,
i \in {1,2, ... I} \\
\end{equation}

Then the result vectors of each word are fed to LSTM in sequence and use the vector of the last word $R_I$ to represent the whole-sentence reminder. 
\begin{equation}\label{eq:2}
R_i = LSTM(M_i) ,
i \in {1,2, ... I} \\    
\end{equation}

With this algorithm, a robot combines the meaning of a single word and the context of the whole reminder to help it to extract attention-related patterns from human reminders.

\subsection{Locating Robot Attention in Visual Perceiving}
The moment the human raises a reminder is when the robots show visually observable abnormal executions. The robot records a video at this specific moment from its own perspective for describing the abnormal executions. The visual features of robot abnormal executions are extracted by the following method.

Each frame of the video is turned into a $448\times448$ size raw image $I$. The images then are converted into a $14\times14\times512$ feature map $V_f$ by a Convolutional Neural Network (CNN) method called VGGnet16\cite{vgg}. The $14\times14$ dimension represents the 196 regions in the $448\times448$ picture and each region denoted by $F_i$ , $i \in [0,195]$ has $32\times32$ pixels. The $512$ is the dimension of the features of each region. In order to combine the word vectors to the image matrix, a perception is used to convert $V_f$ to have the same dimension as the reminder vectors.
\begin{equation}\label{eq:3}
F_{I} = CNN_{VGG}(I)
\end{equation}
\begin{equation}\label{eq:4}
V_{I} = tanh(W_i \cdot F_I + b_i)
\end{equation}

In the Equation \ref{eq:4}, $V_I$ is a matrix. The $i^{th}$ column of $V_I$ stands for the visual feature vector of the $i^{th}$ region of the image.

\subsection{H2R-AT for Attention Transfer}

The \textit{\textbf{H2R-AT}} combines the human reminder and the robot view. By using two layers of attention, the most critical region in the robot view is identified as the actual attention of the robot.

When robots are showing abnormal executions, a robot uses an \textit{\textbf{H2R-AT}} model to gradually filter out unrelated areas within its perceiving scope to focus on the abnormal regions.

We use this stack neural network instead of other method because it had a better alignment between image and natural language, better adapting to dynamic process. With two stacked network, it has a better accuracy than other methods. 

Given the robot visual perceiving feature matrix $V_I$ from the robot vision and the reminder vector $R_I$ from a human supervisor, the robot can reason by the \textit{\textbf{H2R-AT}} model, as shown in Figure \ref{H2R-AT}.

There are two layers in our \textit{\textbf{H2R-AT}} model. In the first layer, a single layer neural network and a softmax function are used to generate the distribution of robot attention to its view.

\begin{equation}\label{eq:5}
h_{1} = tanh((W_{V_I} \cdot V_I) \oplus (W_{R_I} \cdot R_I + b_{R_I}))
\end{equation}
\begin{equation}\label{eq:6}
p_{1} = softmax(W_{p_{1}}\cdot h_{1}+b_{p_{1}})
\end{equation}

 $V_I \in R^{m \times d}$ represents the features of the robot visual perceiving, $m$ represents the dimension of features in a region and $d$ represents the number of regions in robot image perceiving. The vector $R_I\in R^{m}$ represents the reminder features and is a $m$ dimensional vector. Suppose the dimension of $W_{R_I}$ and $W_{V_I}$ is $k \times m$ and the dimension of $W_{h_1}$ is $\textit{1}\times k$, then the matrix $p_1$ is a $d$ dimensional vector and represents the attention distribution of the first layer. $\oplus$ is used to denote the addition between a $m$ dimension vector and a $m \times d$ matrix, which is adding each column of the matrix by the vector.

Then the robot perceiving feature $V_I$ is combined together to a $d$ dimension vector $v$ according to the attention distribution $p_1$ and combines $v$ with $R_I$ to form a vector $u_1$ which has both the information of the robot visual perceiving and the reminder.
\begin{equation}\label{eq:7}
v = p_1 \cdot V_I
\end{equation}
\begin{equation}\label{eq:8}
u_1 = v + R_I
\end{equation}

Because of the use of attention, the more relevant the region is to the abnormal execution, the more likely that a robot will focus on it, which will lead to a more informative $u_1$ and thus a higher accuracy compared to the robot using a full view to reason. However, in a complicated case, one attention layer is not enough to locate the region which is most relevant to the abnormal execution, so the previous attention generating process is iterated by feeding the result of the first attention layer to the second layer, leading to a more fine-grained attention distribution.
\begin{equation}\label{eq:9}
h_{2} = tanh((W_{V'_I} \cdot V'_I) \oplus (W'_{R_I} \cdot u_1 + b'_{R_I}))
\end{equation}
\begin{equation}\label{eq:10}
p_{2} = softmax(W_{p_{2}}\cdot h_{2}+b_{p_{2}})
\end{equation}

Then a new vector $v'$ is generated like $v$ by $p_2$ and added with $u_1$ to generate a more feature distinctive vector $u_2$ which also has both the visual information and the information from the reminder.
\begin{equation}\label{eq:11}
v' = p_2 \cdot V'_I
\end{equation}
\begin{equation}\label{eq:12}
u_2 = v' + u_1
\end{equation}

The generated $u_2$ is used to infer which kind of abnormal execution the robot is making.
\begin{equation}\label{eq:13}
p_{ans} = softmax(W_{u}\cdot u_{2} + b_{u} )
\end{equation}

\subsection{Attention Supported Failure Avoidance}

Based on this research, a correction mechanism is supported. When the abnormal actions detected by the H2R-AT, the correct actions for failure avoidance will be recommended to improve the robot performance.

\begin{equation}\label{eq:14}
    \hat{\alpha} = \mathop{\arg\max}_{\alpha}P(\alpha_i | \alpha_{attention}),
    i \in {1,2, ...} \\
\end{equation}

\section{Validation}

The effectiveness of the \textit{\textbf{H2R-AT}} model was evaluated by both its accuracy and reliability in transferring human attention for robot failure avoidance. The performance of the model is evaluated by comparing the human attention distribution and the model attention distribution.

\begin{figure*}[!ht]
  \centering
 \includegraphics [scale=0.45 ]{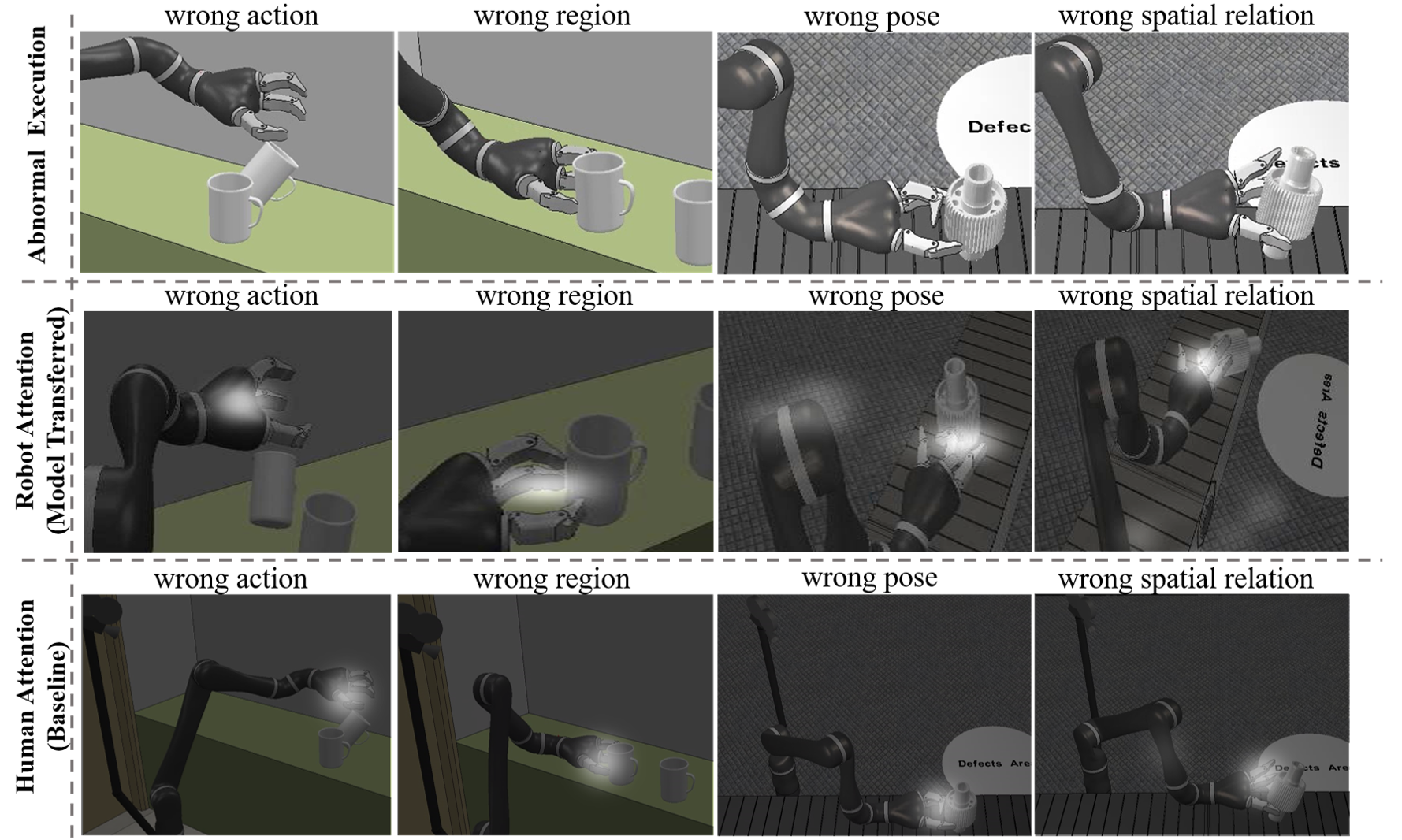}     
 \caption{Visualization of the attention transfer. The \textbf{Baseline} is human attention generated by user study. The three lines in this figure show the simulated robot abnormal execution, robot attention (recommended by \textbf{\textit{H2R-AT}}), and human attention (baseline) respectively. For all four cases, robot attention is highly consistent with human attention, validating the accuracy of \textbf{\textit{H2R-AT}}. Due to the model uncertainties and vague human descriptions, robot attention is slightly more sparse than human attention.}
  \label{visualization}
  \vspace{-1em}
\end{figure*}

\subsection{Experiment Settings: Robot Task Scenarios and Human User Study}

\textbf{Robot Task Design}. To learn and validate the effectiveness of the \textit{\textbf{H2R-AT}} model in guiding robot failure avoidance, two representative task scenarios, ``serve water for a human in a kitchen" and ``pick up a defective gear in a factory" were designed. To represent typical failures in robot executions, four types of basic abnormal robot executions were designed as ``wrong action, wrong pose, wrong region, and wrong spatial relation". ALl other robot executions can be conbined by this four executions. Task scenarios were designed with a JACO robot arm mounted with an HD camera by using the simulation platform CRAIhri, which is developed based on the open-access software V-REP \cite{p16}, and is a widely-used simulation platform in robotics research \cite{p1}\cite{p2}. 
 
In our experiment, a robot arm JACO completed tasks while monitored by a human instructor. The instructor was asked to give verbal reminders to alert the robot when the robot showed abnormal executions. At the moment the human sends alerts, the visual observation from robot perspective was recorded as video training samples. By using the \textbf{\textit{H2R-AT}} model to align both robot visual perceiving and human verbal alerts at the moment robots showed abnormal executions, the nonlinear relation between human attention and robot attention was modeled to guide robot failure avoidance in an early stage. The robot perceiving was recorded from the mounted HD camera.

\textbf{Human User Study}. To learn and validate the \textit{\textbf{H2R-AT}} model, a human user study was conducted to collect verbal instructions for abnormal execution description and suggestions for robot execution corrections. The user study was conducted on the crowd-sourcing platform, Amazon Mechanical Turk \cite{c70}. In total, $252$ English-speaking volunteers were recruited with $1.5$ dollar payment each. They were required to watch a 10 second video containing abnormal executions, and to provide abnormality descriptions, correction suggestion, and the area they paid most attention to, at the moment of detecting abnormal execution. Here in the user study, since the volunteers were asked to give abnormality descriptions, they must pay their attention to the most suspicious area that showed the robot abnormal behavior. Thus, by collecting the regions where they were paying attention to,  the user attention distribution which works as the evaluation baseline of robot attention were generated. 

After filtering the questionnaires, about $12000$ verbal reminders were collected to label $12000$ most-typical images of abnormal robot executions. 

We divided the collected data into two parts equally, one for training and one for testing, these two parts all have four different basic abnormal execution types and two different scenarios. 



\subsection{H2R-AT Performance in Attention Transfer}

\textbf{H2R-AT Model Accuracy}. As shown in Figure \ref{visualization},  human attention was successfully transferred to robot attention. The three lines denote four types of abnormal executions, model-transferred robot attention, and actual human attention (baseline), respectively. 
The accuracy of the \textit{\textbf{H2R-AT}} model in attention transfer is calculated by the average of the precision and recall. Various levels of confidence scores for the predictions made by the \textit{\textbf{H2R-AT}} model were used as the threshold to accept or reject the true positives. On each curve, one dot denotes a recall-precision pair given one threshold; one curve denotes the prediction performance of \textit{\textbf{H2R-AT}} in predicting one category of task scenarios. By setting the confidence threshold as 0.5 in reference, the average precision was about $73.73\%$. The average recall is about $73.63\%$. The precision of each case is $76.73\%$, $72.46\%$, $73.77\%$, and $71.99\%$. The recall of each case is $70.07\%$, $78.75\%$, $70.90\%$, and $74.79\%$. The stable performance and the P-R curves which are close to the upper-right corner show the effectiveness of the \textit{\textbf{H2R-AT}} model in transferring human attention into a robot in various scenarios.

\subsection{H2R-AT Reliability Analysis}

\textbf{Definition of Reliability.} Unlike human attention, which concentrates on one area, robot attention is distributed in several regions due to the model and data uncertainty.
For example, in some ``wrong pose" case, 
the robot mapped some of its attention on the elbow, while correct attention was on the fingers. The most popular attention regions selected by volunteers in the user study were set as the baseline to measure the model reliability of attention transfer. If the model recommended attention regions are inconsistent with the human attention region, then it means predicted attention is unreliable in supporting robot failure avoidance.


\textbf{Analysis of Model Instability.} Vague descriptions caused false attention mapping. The robot attention focused on a same undesired region consistently because the description was not clear enough to point out the exact part causing the error.


There are cases in which the generated robot attention focused on random part which are unrelated to the robot. These kinds of cases mislead the robot to an undesired result. A similarity in features of different parts of the robot perceiving makes these cases unavoidable by simply retraining the model.
Thus, in order to reduce attention focusing on robot-unrelated things, a filter is designed to filter out these distractions, such as wall and carpet. The critical part often appears near the robot and the center of the perceiving. So the edge parts of the robot view were filtered out to help the robot with accurate attention mapping. It turns out that with the filter, the generated attention is more focused on the robot and the object.


\section{Conclusion}
In this paper, \textit{\textbf{H2R-AT}}, a novel model using human attention to avoid robot execution failure was proposed. The robot was enabled to identify its abnormal executions by interpreting human verbal reminders. Four types of robot abnormal executions - wrong action, wrong region, wrong pose, and wrong spatial relation - in both daily and industrial scenarios were designed. Volunteers were recruited to provide verbal reminders for the robots and labeled their concerned executions for training the \textit{\textbf{H2R-AT}}. With an average accuracy of $73.68\%$ in transferring human attention and $67.04\%$ performance improvement, the feasibility of verbally transferring human attention to robots for failure avoidance was validated, showing the great potential in using this \textit{\textbf{H2R-AT}} model for naturally integrating human intelligence for robot failure avoidance, in scenarios from daily assistance to cooperative manufacturing. 


Though this work was based on simulated environment, there are enough facts proving that this model can be used in the real-world environment. To start with, in this work, robot executions were the only part that was simulated, the model itself and the instructions we collected from the volunteers were real. That means, without considering mechanical failures, the environment was no different from the real-world environment. Moreover, the four tasks that were validated in this work covered the four basic robot abnormal actions. Thus, every robot abnormal behavior in the real-world is essentially one of the tasks in this work. 

To implement this model to practically guide robot executions in a real-world environment, data of the real world, i.e. appropriate visual observations of practical robot behaviors, as well as human verbal descriptions for the abnormal robot executions need to be provided to train a practical model for guiding real-world human-robot interaction. In the future, novel attention-based correction methods will be designed to accurately correct robot executions after human reminders. Also, the attention region identification can be improved by using some rule-based methods to narrow down the searching.

\addtolength{\textheight}{-4cm}   


\balance

\end{document}